\documentclass[fleqn,10pt]{olplainarticle}
\usepackage[utf8]{inputenc} 
\usepackage[T1]{fontenc}    
\usepackage{hyperref}       
\usepackage{url}            
\usepackage{booktabs}       
\usepackage{nicefrac}       
\usepackage{microtype}      
\usepackage{mydef}
\usepackage{color}
\usepackage{wrapfig,lipsum,booktabs}
\usepackage{microtype}
\usepackage{graphicx}
\usepackage{subfigure}
\usepackage{booktabs} 
\usepackage{amsfonts,amsmath}
\usepackage{multirow}
\usepackage{enumitem}

\usepackage{algorithm}
\usepackage{algpseudocode}

\usepackage[percent]{overpic}
\usepackage[mathscr]{euscript}
\newcommand{\Ti}{\mathcal{T}_i}
\newcommand{\Ttr}{\mathcal{T}_{tr}}
\newcommand{\Tval}{\mathcal{T}_{val}}
\usepackage{mathtools}

\title{A Meta-Learning Framework for Generalized Zero-Shot Learning}

\author[1]{Vinay Kumar Verma}
\author[1]{Dhanajit Brahma}
\author[1]{Piyush Rai}
\affil[1]{Department of CSE\\Indian Institute of Technology Kanpur}
\affil[1]{\{vkverma,dhanajit,piyush\}@cse.iitk.ac.in}


	\begin{abstract}
    		Learning to classify unseen class samples at test time is popularly referred to as zero-shot learning (ZSL). If test samples can be from training (seen) as well as unseen classes, it is a more challenging problem due to the existence of strong bias towards seen classes. This problem is generally known as \emph{generalized} zero-shot learning (GZSL). Thanks to the recent advances in generative models such as VAEs and GANs, sample synthesis based approaches have gained considerable attention for solving this problem. These approaches are able to handle the problem of class bias by synthesizing unseen class samples. However, these ZSL/GZSL models suffer due to the following key limitations: $(i)$ Their training stage learns a class-conditioned generator using only \emph{seen} class data and the training stage does not \emph{explicitly} learn to generate the unseen class samples; $(ii)$ They do not learn a generic optimal parameter which can easily generalize for both seen and unseen class generation; and $(iii)$ If we only have access to a very few samples per seen class, these models tend to perform poorly. In this paper, we propose a meta-learning based generative model that naturally handles these limitations. The proposed model is based on integrating model-agnostic meta learning with a Wasserstein GAN (WGAN) to handle $(i)$ and $(iii)$, and uses a novel task distribution to handle $(ii)$. Our proposed model yields significant improvements on standard ZSL as well as more challenging GZSL setting. In ZSL setting, our model yields 4.5\%, 6.0\%, 9.8\%, and 27.9\% relative improvements over the current state-of-the-art on CUB, AWA1, AWA2, and aPY datasets, respectively.
    	\end{abstract}
    	\vspace{-10pt}

\begin{document}

\flushbottom
\maketitle
\thispagestyle{empty}

    	\section{Introduction}
    	\label{submission}
    	With the ever-growing quantities, diversity, and complexity of real-world data, machine learning algorithms are increasingly faced with challenges that are not adequately addressed by traditional learning paradigms. For classification problems, one such challenging setting is where test-time requires correctly labeling objects that could be from classes that were not present at training time. This setting is popularly known as Zero-Shot Learning (ZSL), and has drawn a considerable interest  recently~\cite{cmt,conse,verma2017simple,changpinyo2016synthesized,ESZSL2015,xian2018feature,vermageneralized,calibration_generalized,romera2015embarrassingly,Chen_2018_CVPR,mishra2017generative,bucher2017generating,Zero-ShotTaskTransfer,cycle-consistancy,saligram_cvpr19,cada-vae,kumar2019generative}. ZSL algorithms typically rely on class-descriptions (e.g., human-provided class attribute vectors, textual description, or word2vec embedding of class name). These class-description/class-attributes are leveraged to transfer the knowledge from \emph{seen} classes (i.e., classes that were present at training-time) to \emph{unseen} classes (i.e., classes only encountered in test data).
    	
    	Driven by the recent advances in generative modeling \cite{wgan,VAE,progressivekarras}, there is a growing interest in generative models for ZSL. Broadly, these models learn to generate/synthesize ``artificial'' examples from unseen classes~\cite{vermageneralized,cycle-consistancy,xian2018feature,mishra2017generative,lisgan,cada-vae,khare2019generative}, conditioning on their class attributes, and learn a classifier using these synthesized examples. Despite the recent progress on such approaches, these still have some key limitations. Firstly, while the goal of these approaches is to generate unseen/novel class examples given the respective class attributes, these models are trained using data (inputs and the respective class attributes) from the seen classes~\cite{vermageneralized,xian2018feature,cycle-consistancy,lisgan} and do not explicitly learn to generate the unseen class samples during training.
    	Consequently, these generative ZSL models show a large quality gap between the synthesized unseen class inputs and actual unseen class input. To mimic the ZSL setting explicitly, we propose a novel variant of the standard meta-learning based approach~\cite{finn2017model}. Notably, in our variant, the meta-train and meta-validation classes are \emph{disjoint}.
    	
    	The second limitation of existing ZSL/GZSL models is that they do not learn an optimal parameter which can easily generalize to the seen/unseen class generation. Our meta-learning framework learns such an optimal parameter that can quickly adapt to the novel classes (meta-test) with few gradient steps. \cite{prototypical,matching} show that even with the \emph{zero-gradient step} (without fine-tuning),
    	meta-learning learns to generalize novel class samples/task. We build on this idea to train a class-conditioned WGAN for sample generation.  
    	
    	The third key limitation is that all the existing ZSL methods rely on the availability of a significant number of labeled samples from each of the seen classes. This itself is a severe requirement and may not be met in practice (e.g., we may only have a handful, say 5, or 10 examples from each seen class). Note that this setting is somewhat similar to few-shot learning or meta-learning~\cite{finn2017model} where the goal is to learn a classifier using very few examples per class, but all the test/unseen class are assumed to have few samples in test time. In contrast, in ZSL, we do not have any labeled training data from unseen classes. Our meta-learning based formulation is naturally suited to this setting where only a few samples per class are available.
    	
    	Our approach is primarily based on learning a generative model that can synthesize inputs from any class (seen/unseen), given the respective class-attributes/description. However, unlike recent works on synthesis based ZSL models~\cite{lisgan,noisy_text,vermageneralized,xian2018feature,cycle-consistancy}, we endow the generator the capability to meta-learn using very few examples per seen class. To this end, we develop a meta-learning based \emph{conditional} Wasserstein GAN~\cite{wgan} (conditioning on the class-attributes) which has a generator and a discriminator modules augmented with a classifier. Each module is associated with a meta-learning agent, to facilitate learning with a very small number of seen class inputs. Also, the novel task distribution helps to mimic the ZSL behavior, i.e., the generative model not only learns to generate the seen class samples but the unseen class samples as well. We would also like to highlight that, although we develop this model with the focus being ZSL and generalized ZSL, our ideas can be used for the task of supervised few-shot generation~\cite{few-shotimage}, which is the problem of learning to generate data given very few examples to learn the data distribution. Our main contributions are summarized below:
    	\vspace{-2pt}
    	\begin{itemize}[leftmargin=0.35cm]
    		\item We develop a novel meta-learning framework for ZSL and generalized ZSL by learning to synthesize examples from unseen classes, given the respective class-attributes. Notably, our framework is based on model-agnostic meta-learning~\cite{finn2017model}, which enables the synthesis of high-quality examples. This helps to overcome the above mentioned second and third limitation.
    		\item We propose a novel episodic training for the meta-learning based ZSL where, in each episode, the \emph{training-set} and \emph{validation-set} classes are disjoint.
    		This helps \emph{learning to generate the novel class examples in training itself}. This contributes in overcoming the above mentioned first limitation.
    	\end{itemize}
    	\vspace{-10pt}
    	\section{Notation, Preliminaries, Problem Setup}
    	A typical ZSL setting is as follows: We have $S$ \emph{seen} classes with labelled training data and $U$ \emph{unseen} classes with no labelled data present during the training time. The test data can be either exclusively from unseen classes (standard ZSL setting), or can be from both unseen and seen classes (\emph{generalized} ZSL setting). We further assume that we are provided \emph{class-attribute vectors} for the seen as well as unseen classes $\mathbf{A} =\{\mathbf{a_c}\}_{c=1}^{S+U}$, where $\mathbf{a_c} \in \mathbb{R}^d$ is the class-attribute vector of class $c$. These class-attribute vectors are leveraged by the ZSL algorithms to transfer the knowledge from seen to unseen classes.
    	
    	Existing ZSL algorithms 1assume that we have access to a significant number of examples from each of the seen classes. This may however not be the case; in practice, we may have very few examples from each of the seen classes. We train our model in $N$-way $K$-shot setting such that it can handle the ZSL problem when only very few samples are available per seen class. We choose the model-agnostic meta learning (MAML)~\cite{finn2017model} as our meta-learner due to its generic nature; it only requires a differentiable model and can work with any loss function. 
    	\begin{figure*}[!htbp]
    		\centering
    		\includegraphics[scale=0.43]{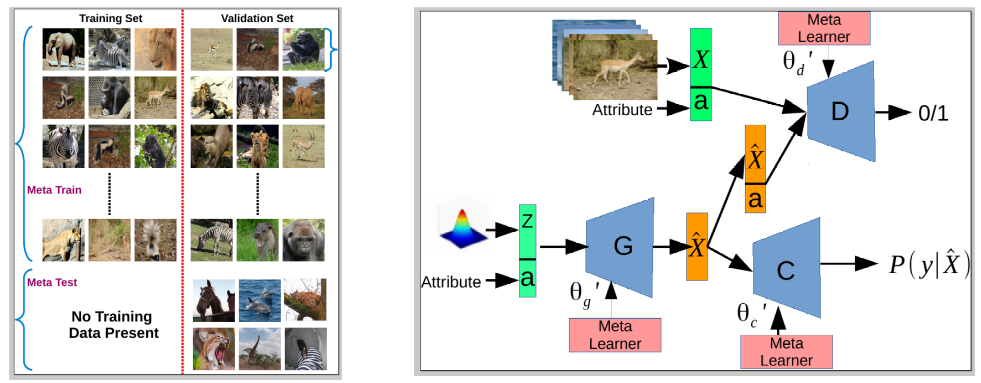}
    		\vspace{-5pt}
    		\caption{\textbf{Left:} Task episode for zero-shot meta-learning. For each task $\Ti=\{\Ttr,\Tval\}$, \emph{training set} $\Ttr$ and \emph{validation set} $\Tval$ classes are disjoint. In the ZSL setup, we have \emph{zero} training examples from the \emph{meta-test} set. \textbf{Right:} The proposed architecture model. ${X}$: ResNet-101 feature vector.}
    		\label{fig:main_image}
    		\vspace{-10pt}
    	\end{figure*}
    	\vspace{-5pt}
    	\subsection{Model-Agnostic Meta-Learning (MAML)}
    	\label{sec:maml}
    	MAML~\cite{finn2017model} is an optimization based meta-learning framework designed for few-shot learning. The model is designed in such a way that it can quickly adapt to a new task with the help of only few training examples. MAML assumes that model $f_\theta$ is parameterized by learnable parameters $\theta$ and the loss function is smooth in $\theta$ that can be used for the gradient-descent based updates.
    	
    	Let $p(\mathcal{T})$ be the distribution of tasks over the \emph{meta-train} set. MAML defines the notion of a ``task'' such that a task $\Ti \sim p(\mathcal{T})$ represents a set of labeled examples and MAML splits this set further into a training set $\Ttr$ and a validation set $\Tval$, i.e., $\Ti=\{\Ttr,\Tval\}$. The split is done such that $\Ttr$ has very few examples per class. 
    	We follow the general notion of $N$-way $K$-shot problem~\cite{matching}
    	, i.e., $\Ttr$ contains $N$ classes with $K$ examples from each class.
    	The model is trained using an episodic formulation where each round samples a batch of tasks and uses gradient-descent based updates (inner loop) for the parameters $\theta_i$ specific to each task $\Ti$.
    	The meta-update step (outer loop) then aggregates the information from all these ``local'' updates to update the overall model parameters $\theta$, using gradient descent update.
    	
    	For task $\Ti$, its local parameters $\theta_i$ are updated by starting with the global model parameters $\theta$, and using a few gradient based updates computed on $\Ttr$ from task $\Ti$. Assuming a single step of update, this can be written as: $\theta_i'=\theta-\alpha\nabla_\theta \Lcal_{\Ttr}(f_\theta)$.Here, $\alpha$ is the hyper-parameter and $\Lcal$ denotes the loss function being used. The overall global/meta objective defined over the multiple tasks sampled from task distribution $p(\mathcal{T})$ can be defined as:
    	\begin{equation}
    	\small
    	\sum_{\Ti\sim p(\mathcal{T})} \Lcal_{\Ttr}(f_{\theta_i'})= \sum_{\Ti\sim p(\mathcal{T})} \Lcal_{\Ttr}(f_{\theta-\alpha\Lcal_{\Ttr}(f_\theta)})
    	\label{eq:maml-full}
    	\end{equation}
    	
    	Assuming a gradient descent based optimization of the global objective in Eq.~\ref{eq:maml-full}, a single-step gradient descent update for the global parameter can be written as: $\theta \leftarrow \theta-\beta \nabla_\theta \sum_{\Ti\sim p(\mathcal{T})} \Lcal_{\Tval}(f_{\theta_i'})$.
    	\subsection{Zero-Shot Meta-Learning (ZSML)}
    	\label{sec:zsmaml}
\begin{figure}[!ht]
    		\vspace{-10pt}
    		\centering
    		\includegraphics[scale=0.6]{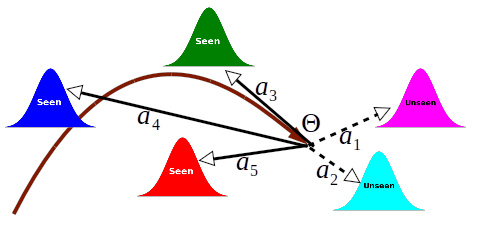}
    		\vspace{-10pt}
    		\caption{Our proposed ZSML learns a generic optimal parameter $\Theta$ such that it can generate the seen/unseen class samples with zero-gradient step update  conditioned on the class attribute $a_i$ (at test time).}
    		\label{fig:zsml_theta}
    		\vspace{-10pt}
    	\end{figure}

    	The meta-learning framework \cite{finn2017model,ravi2016optimization,matching,prototypical} can quickly adapt to a new task with the help of only a few gradient steps.
    	The quick adaption is only possible for the model if it learns the optimal parameter $\theta$ in the parameter space that is unbiased towards the meta-train data.
    	The learned parameters are close to the optimal parameters for both meta-train and meta-test data respectively (as shown in Figure~\ref{fig:zsml_theta}).
    	It is already demonstrated in \cite{matching,prototypical} where without fine-tuning (using \emph{zero} gradient steps, i.e., not making any update) on the meta-test, the meta-learning model shows better/similar performance.
    	Our ZSML approach is primarily motivated by high-quality generalization ability of the meta-learning towards the seen/unseen class samples.
    	We use the meta-learning framework to train a generative adversarial network conditioned on class attributes, that can generate the novel class samples.
    	A key difference with MAML, to \emph{mimic} the \emph{ZSL behaviour}, is that for each task $\Ti=\{\Ttr,\Tval\}$, the classes of $\Ttr$ and $\Tval$ are disjoint, whereas, in MAML, both set of classes are the same.
    	Therefore, the training is done in such a way that $\Ttr$ acts as \emph{seen} classes and $\Tval$ acts as \emph{unseen} classes.
    	The inner loop of the meta-learning optimizes the parameters using $\Ttr$, and final parameters are updated over the loss of the $\Tval$ (containing disjoint set of classes).
    	Therefore, the model \emph{learns to generate the novel class during the training itself}. In the next section, we describe our complete model (shown in Figure~\ref{fig:main_image} (right)).
    	
    	\vspace{-10pt}
    	\section{Meta-Learning based Adversarial Generation}
    	\label{sec:zsadv}
    	The core of our ZSL model (Figure~\ref{fig:main_image} right) is a generative adversarial network \cite{GAN}, coupled with (1) an additional classifier module trained to correctly classify the examples generated by the generator module; and (2) meta-learners in each of the three modules (Generator ($G$), Discriminator ($D$), and Classifier ($C$)). We use the Wasserstein GAN \cite{wgan} architecture due to its nice stability properties. We assume $\theta_d$, $\theta_g$ and $\theta_c$ to be the parameters of the Discriminator, Generator and Classifier, respectively. 
    	
    	Our model follows the episode-wise training akin to MAML (however, $\Ttr$ and $\Tval$ classes are disjoint in our ZSL setting). There are three meta-learners in the model, one for each $D$, $G$ and $C$, but $G$ and $C$ are optimized jointly. From now on, we will denote the parameters for $G$ and $C$ as a joint set of parameters $\theta_{gc}=[\theta_g,\theta_c]$.
    	
    	For each task $\Ti=\{\Ttr,\Tval\}$, sampled from the task distribution $p(\mathcal{T})$, $\Ttr$ is used by the meta-learners (in the inner loop) of $D$ and $G$. $\Tval$ is used to calculate the loss over the most recent parameters of the meta-learners. For our model, the generator network $G: {\mathbf{Z}}\times {\mathbf{A}} \rightarrow {\mathbf{\hat{X}}}$ takes input as, a random noise $\mathbf{z}\sim \mathbf{\mathcal{N}(0,I)}$ ($\mathbf{z}\in \mathbf{Z}$), concatenated with the class-attribute vector $\mathbf{a_c}$ of a class. $G$ produces a sample $\mathbf{\hat{x}} \in \mathbf{\hat{X}}$ that is similar to a real sample from \emph{that} class. The discriminator network $D: \mathbf{X}\times \mathbf{A}\rightarrow{[0,1]}$ tries to distinguish such generated samples (concatenated with attributes) from the actual sample $\mathbf{X}$ (real data distribution).
    	In addition, the goal of the classifier network $C:\mathbf{\hat{X}}\rightarrow{\mathcal{Y}}$ is to take the generated sample $\mathbf{\hat{x}}$ from $G$ and classify it into the original class $c\in \mathcal{Y}$ where $\mathcal{Y}$ is the set of both \emph{seen} and \emph{unseen} classes. Presence of the classifier module $C$ ensures that the generated sample has the same characteristics as that of samples from that class. 
    	
    	We now describe the objective function of our model. Let $\Lcal_{\Ti}^{D}$
    	denote the meta-learner objective of the discriminator $D$ and $\Lcal_{\Ti}^{GC}$ denote the meta-learner objective of the generator $G$ and the classifier $C$, on the task $\Ti$. The meta-learner objective $\Lcal_{\Ti}^{D}$ for discriminator $D$ can be defined as:
    	\begin{equation} \label{eq:4}
    	\small
    	\Lcal_{\Ti}^{D}(\theta_d)=\mathbb{E}_{\Ti} D(\mathbf{x},\mathbf{a_c}|\theta_d) - \mathbb{E}_{\mathbf{a_c},\mathbf{\hat{x}}\sim P_{\theta_g}} D(\mathbf{\hat{x}},\mathbf{a_c}|\theta_d)
    	\end{equation}
    	Here, $\mathbf{a_c}\in \mathbf{A}$ is attribute vector of samples belonging to $\Ti$. The objective in Eq.~\ref{eq:4} (to be maximized) essentially says that the discriminator should have $D(.)$ large for real examples and small for generated examples. The meta-learner objective $\Lcal_{\Ti}^{GC}$ for generator $G$ and classifier $C$ is given as:
    	\begin{equation}\label{eq:5}
    	\small
    	\begin{split}
    	\Lcal_{\Ti}^{GC}(\theta_{gc})= & - \mathbb{E}_{\mathbf{a_c},\mathbf{{z}}\sim \mathcal{N}(0,I)} D({G(\mathbf{a_c},\mathbf{z}|\theta_g)},\mathbf{a_c}|\theta_d) +C(y|\hat{\mathbf{x}},\theta_c)
    	\end{split}
    	\end{equation}
    	This objective (to be minimized) says that the generator's output $G(\mathbf{a_c},\mathbf{z}|\theta_g)$ should be such that $D(.)$ is large, as well as the classifier's loss $C$ should be small (i.e., the classifier should predict the correct class for generated example $\mathbf{\hat{x}}$).
    	Having defined the individual objectives, the overall objective for the meta-learner (inner loop) update for task ${\Ti}$:
    	\begin{equation}
    	l_{\Ti}^D=\max_{\theta_d}\Lcal_{\Ti}^{D}(\theta_d) \quad \text{and} \quad
    	l_{\Ti}^{GC}=\min_{\theta_{gc}} \Lcal_{\Ti}^{GC}(\theta_{gc})
    	\end{equation}
    	The meta-learner gradient ascent update for the discriminator over a task $\Ti$ will be:
    	\begin{equation}
    	\theta_d'=\theta_d+ \eta_1 \nabla_{\theta_d} l_{\Ttr \in \Ti}^D(\theta_{d})
    	\end{equation}
    	Similarly the meta-learner gradient descent update for the generator and classifier over $\Ti$ will be:
    	\begin{equation}
    	\theta_{gc}'=\theta_{gc}- \eta_2 \nabla_{\theta_{gc}} l_{\Ttr \in \Ti}^{GC}(\theta_{gc})
    	\end{equation}
    	The model parameters are learned by optimizing Eq~\ref{eq:4} and Eq~\ref{eq:5} over a batch of sampled tasks from the task distribution $p(\mathcal{T})$. The overall meta-objective for the discriminator and generator is:
    	\vspace{-4pt}
    	\begin{equation} \label{eq:7}
    	\small
    	\theta_d'=\theta_d+ \eta_1 \nabla_{\theta_d} \sum_{ \Ttr \in \Ti \sim p(\mathcal{T})}l^D_{\Ttr}(\theta_{d})
    	\end{equation}
    	\begin{equation}\label{eq:8}
    	\theta_{gc}'=\theta_{gc}- \eta_2 \nabla_{\theta_{gc}} \sum_{ \Ttr \in \Ti \sim p(\mathcal{T})}l^{GC}_{\Ttr}(\theta_{gc})
    	\end{equation}
    	Unlike to standard MAML in the inner loop (i.e. Eq:\ref{eq:7} and \ref{eq:8}) are optimize on the set of task instead of per task. We observe that this increase the stability of the WGAN training. Having meta-learned the discriminator parameters from the meta-training phase (performed using the seen class examples), the discriminator's objective function w.r.t. the unseen class examples in the validation meta-set is given by:
    	\begin{align} \label{eq:12}
    	\small
    	\begin{split}
    	&\max_{\theta_d}\sum_{ \Tval \in \Ti \sim p(\mathcal{T})}l^D_{\Tval}(\theta_{d}')\\
    	=&\max_{\theta_d} \sum_{ \Tval \in \Ti \sim p(\mathcal{T})}l^D_{\Tval}(\theta_d+ \eta_1 \nabla_\theta l^D_{\Ttr}(\theta_{d}))  
    	\end{split}
    	\end{align}
    	Therefore, the final update of the discriminator $D$ for the batch is:
    	\begin{equation}
    	\small
    	\theta_d\leftarrow\theta_d+ \beta_1 \nabla_{\theta_d} \sum_{ \Tval \in \Ti \sim p(\mathcal{T})}l^D_{\Tval}(\theta_{d}')
    	\end{equation}
    	Here, $\beta_1$ is the learning rate for the meta-step and $\theta_d'$ is the optimal parameter provided by the inner loop of meta-learner for the discriminator. Likewise, the generator's and classifier's objective function w.r.t. the unseen class examples in $\Tval$ is given by:
    	\begin{align} \label{eq:14}
    	\small
    	\begin{split}
    	&\min_{\theta_{gc}}\sum_{ \Tval \in \Ti \sim p(\mathcal{T})}l^{GC}_{\Tval}(\theta_{gc}')\\
    	\xRightarrow{Update} \theta_{gc}&\leftarrow\theta_{gc}- \beta_2 \nabla_{\theta_{gc}} \sum_{ \Tval \in \Ti \sim p(\mathcal{T})} l^{GC}_{\Tval}(\theta_{gc}')
    	\end{split}
    	\end{align}
    	Eq.~\ref{eq:14} performs the meta-optimization across the batch of task for the generator and classifier. Again, note that, each task $\Ti=\{\Ttr,\Tval\}$ is partitioned into \emph{training set} $\Ttr$ and \emph{validation set} $\Tval$, such that the classes are disjoint. In contrast, traditional meta-learning \cite{finn2017model} designed for few-shot learning assumes that the set of classes in $\Tval$ is same as the set of classes in $\Ttr$. This disjoint setup for $\Ttr$ and $\Tval$ is designed for zero-shot learning in order to mimic the problem setting which requires predicting the labels for examples from unseen classes not present at training time.
    	\vspace{-5pt}
    	\subsection{Example Generation and Zero-Shot Classification}
    	\label{sec:gen_classification}
    	After training the model, we can generate the unseen class examples given the respective class-attribute vectors. The generation of the novel class examples is done as:
    	\begin{equation}
    	\small
    	\mathbf{\hat{x}}=G_{\theta_g}(\mathbf{z},\mathbf{a}_c): \quad \mathbf{a}_c \in \mathbb{R}^d,  c \in \{S+1, \dots S+U\}
    	\end{equation}
    	Here, $\mathbf{z}\sim \mathcal{N}(0,\Imat)$ and $\mathbf{z}\in \mathbb{R}^k$. Once we have generated samples from the unseen classes, we can train any classifier (e.g., SVM or softmax classifier) with these samples as labeled training data. In \emph{generalized} ZSL setting, we synthesize samples from both seen and unseen class. We use the unseen class generated samples and actual/generated examples from seen classes to train a classifier with the label space being the union of seen and unseen classes. In practice, we found that using generated samples from seen classes (as opposed to actual samples) tends to perform better in the generalized ZSL setting. A justification for this is that the generated sample quality is uniform across seen and unseen class examples. 
    	
    	\begin{table*}[t]
    		\small
    		\centering
    		\scalebox{0.95}{
    			\addtolength{\tabcolsep}{12.0pt}
    			\begin{tabular}{|l| c | c | c | c |c|} 
    				\hline
    				\textbf{Method}& {\textbf{SUN}} & {\textbf{CUB}} & {\textbf{AWA1}} & {\textbf{AWA2}}  & \textbf{aPY}\\ \hline
    				\hline
    				\textbf{LATEM} \cite{latem}  & 55.3  & 49.3  & 55.1  & 55.8 & 35.2\\
    				\textbf{SJE} \cite{SJE}  & 53.7  & 53.9  & 65.6  & 61.9  &32.9\\
    				\textbf{ESZSL} \cite{ESZSL2015}  & 54.5  & 53.9  & 58.2  & 58.6  & 38.3\\
    				\textbf{SYNC}\cite{changpinyo2016synthesized}  & 56.3  & 55.6  & 54.0 & 46.6 &23.9\\
    				\textbf{SAE} \cite{SAE2017} & 40.3  & 33.3  & 53.0 & 54.1 &8.3 \\
    				\textbf{DEM} \cite{dem} & 61.9 & 51.7 & 68.4& 67.1 & 35.0\\ 
    				\textbf{DCN} \cite{calibration_generalized} & 61.8 & 56.2 & --& 65.2 & 43.6\\ 
    				\textbf{ZSKL} \cite{zskl} & 61.7 & 51.7 & 70.1 & 70.5 & 45.3\\
    				\hline
    				\hline
    				\textbf{GFZSL}\cite{verma2017simple}  & 62.6   & 49.2  & 69.4 & 67.0 & 38.4 \\
    				\textbf{SP-AEN} \cite{Chen_2018_CVPR}  & -- & 55.4 & -- & 58.5 & 24.1\\
    				\textbf{CVAE-ZSL}\cite{mishra2017generative}  & 61.7  & 52.1  & {71.4} & 65.8  &--\\
    				\textbf{cycle-UWGAN} \cite{cycle-consistancy} & 59.9 & 58.6 &-- & 66.8  &--\\
    				\textbf{f-CLSWGAN} \cite{xian2018feature} & 60.8 & 57.3 & -- & 68.2 &--\\
    				\textbf{SE-ZSL} \cite{vermageneralized} & \textbf{63.4} &{59.6} & {69.5} &  {69.2}&--\\
    				\textbf{VSE-S} \cite{saligram_cvpr19} & -- & 66.7 & -- &  {69.1}& 50.1\\
    				\textbf{LisGAN} \cite{lisgan} & 61.7 & 58.8 & -- &  70.6 & 43.1\\
    				\hline\hline
    				\textbf{ZSML} Softmax (Ours) & 60.2 & \textbf{69.6} & \textbf{73.5} & \textbf{76.1}  & \textbf{64.1}\\
    				\textbf{ZSML} SVM (Ours) & 60.1 & \textbf{69.7} & \textbf{74.3} & \textbf{77.5}  & \textbf{64.0}\\
    				\hline
    			\end{tabular}
    		}
    		\caption{ZSL result using the per-class mean metric \cite{xian2018zero}. The non-generative models are mentioned at the top and the generative models are mentioned at the bottom. All compared methods use CNN-RNN feature for CUB dataset.}
    		\label{tab:zsl}
    		\vspace{-15pt}
    	\end{table*}
    	\vspace{-10pt}
    	\section{Related Work}
    	\label{sec:relwork}
    	Some of the earliest works on ZSL were based on directly or indirectly mapping the inputs to the class-attributes~\cite{IAP,conse,cmt}. The learned mapping is used at inference time, this mapping first projects the unseen data to class-attribute space and then uses nearest neighbour search to predict the class. In a similar vein, other approaches \cite{ESZSL2015,changpinyo2016synthesized} also consider the relationship between seen and unseen classes. They represent the parameters of each unseen class as a similarity weighted combination of the parameters of seen classes. All of these models require plenty of data from the seen classes, and also do not work well in GZSL setting~\cite{vermageneralized,xian2018zero}.
    	
    	Because of the wide applicability and more realistic setting the ZSL framework also applied on the different domain like Zero-Shot Task Transfer \cite{zero-shottask_cvpr19}, zero-shot sketch-based image retrieval \cite{shen2018zero,kumar2019generative}, zero-shot knowledge distillation \cite{nayak2019zero}, zero-shot action recognition \cite{xu2015semantic,gan2015exploring,mishra2018generative,mandal2019out} etc. These fields are not in the scope of this paper. In this paper, we focus on zero-shot image classification. Therefore in the rest of the section, we discuss the ZSL framework for the image classification. Here note that our approach is generic and it can be easily applied over the other ZSL domain also.
    	
    	Another prominent approach for ZSL focuses on learning the bilinear compatibility between the visual space and the semantic space of classes. \cite{akata2013label,frome2013devise,SJE,ESZSL2015,SAE2017} are based on computing a linear/bilinear compatibility function. \cite{sse} embeds the inputs based on the semantic similarity. Some of the ZSL methods assume that all the unseen class inputs are also present at the time of training without the class labels. These \emph{transductive} methods have extra information about all the unlabelled data of the unseen class, which leads to improved predictions as compared to the inductive setting~\cite{song2018transductive,xu2017transductive}. Note that the transductive assumption is not very realistic since often test data is not available at the time of training.
    	
    	The generalized ZSL (GZSL) \cite{vermageneralized,chao2016empirical,xian2018zero,xian2018feature} problem is arguably a very realistic and challenging problem wherein, unlike the ZSL problem, the training (seen) and the test (unseen) classes are not disjoint. Most of the previous models that perform well on standard ZSL fail to handle the biases towards predicting seen classes. 
    	Recently, generative models~\cite{Chen_2018_CVPR,xian2018feature,verma2017simple,guo2017synthesizing,wang2017zero} have shown promising results for both ZSL and GZSL setups. \cite{verma2017simple} used a simple generative model based on the exponential family framework while \cite{guo2017synthesizing} synthesized the classifier weights using class attributes. Recent generative approaches for ZSL are mostly based on VAE \cite{VAE} and GAN \cite{GAN}. Among these,~\cite{vermageneralized,bucher2017generating,fvaegan} are based on the VAE architectures while \cite{xian2018feature,Chen_2018_CVPR,lisgan,cycle-consistancy} use adversarial sample generation based on the class conditioned attribute. The recent approaches based on VAE and GAN show very competitive results. A particular advantage of the generative approaches is that, by using synthesized samples, we can convert the ZSL problem to the conventional supervised learning problem that can handle the biases towards the seen classes. The meta-learning approach are already tried for the ZSL \cite{hu2018correction} to correct the learned network. To the best of our knowledge MAML~\cite{finn2017model} based approach over GAN has not been investigated yet. The meta-learning based adversarial generation model shows significant performance improvement, whereas the recent generative ZSL models have saturated.

    	\vspace{-7pt}
    	\section{Experiments and Results}
    	We perform a comprehensive evaluation of our approach \textbf{ZSML} (\textbf{Z}ero-\textbf{S}hot \textbf{M}eta-\textbf{L}earning) by applying it on both standard ZSL and generalized ZSL problems and compare it with several state-of-the-art methods. We also perform several ablation studies to demonstrate/disentangle the benefits of the various aspects of our proposed approach.\footnote{We will provide the code and data upon publication.} We evaluate our approach on the following benchmark ZSL datasets: SUN \cite{xiao2010sun} and CUB \cite{welinder2010caltech} which are fine-grained and considered very challenging; AWA1~ \cite{lampert2009learning} and AWA2~ \cite{xian2018zero}; aPY~ \cite{farhadi2009describing} with diverse classes that makes this dataset very challenging. For CUB dataset, we use CNN-RNN textual features ~\cite{reed2016learning} as class attributes, similar to the approaches mentioned in Table~\ref{tab:zsl}~and~\ref{tab:gzsl}.
    	Due to the lack of space, the complete \emph{Algorithm} and details about the datasets are provided in the \emph{Supplementary Material}. The generator and discriminator are 2-hidden layer networks with hidden layer size 2048 and 512, respectively. More details of the model architecture, experimental setup and various hyperparameters are provided in the \emph{Supplementary Material}.
    	\begin{table*}[!t]
    		\centering
    		\scalebox{0.9}{
    			\addtolength{\tabcolsep}{-1.2pt}
    			\begin{tabular}{|l|c c c | c c c | c c c | c c c |} 
    				\hline
    				\multirow{2}{*}{\textbf{Method}} & \multicolumn{3}{c|}{\textbf{AWA1}} & \multicolumn{3}{c|}{\textbf{CUB}} & \multicolumn{3}{c|}{\textbf{aPY}} & \multicolumn{3}{c|}{\textbf{AWA2}} \\ 
    				\cline{2-13} &\textbf{U}& \textbf{S} & \textbf{H} & \textbf{U }& \textbf{S } & \textbf{H} & \textbf{U}& \textbf{S} & \textbf{H} & \textbf{U}& \textbf{S} & \textbf{H} \\ 
    				\hline
    				
    				\textbf{SJE} \cite{SJE} & 11.3 &74.6& 19.6& 23.5& 59.2& 33.6& 3.7& 55.7& 6.9& 8.0& 73.9& 14.4\\
    				\textbf{ESZSL} \cite{ESZSL2015} & 6.6 &75.6& 12.1& 12.6& 63.8& 21.0& 2.4& 70.1& 4.6& 5.9& 77.8& 11.0 \\
    				\textbf{SYNC}\cite{changpinyo2016synthesized}& 8.9& 87.3 &16.2& 11.5& {70.9}& 19.8& 7.4& 66.3& 13.3& 10.0& 90.5& 18.0 \\
    				\textbf{SAE} \cite{SAE2017} & 8.8& 18.0& 11.8& 7.8& 54.0& 13.6& 0.4& 80.9& 0.9& 1.1& 82.2& 2.2 \\
    				\textbf{LATEM} \cite{latem}& 7.3& 71.7 &13.3& 15.2& 57.3& 24.0& 0.1& 73.0& 0.2& 11.5& 77.3& 20.0  \\
    				\textbf{DEVISE} \cite{frome2013devise} & 13.4& 68.7& 22.4& 23.8& 53.0& 32.8& 4.9& 76.9& 9.2& 17.1& 74.7& 27.8 \\
    				\textbf{DEM} \cite{dem} & 32.8&    84.7&47.3 & 19.6& 57.9&29.2& 11.1 &75.1 & 19.4 &30.5 &86.4&45.1\\
    				\textbf{ZSKL} \cite{zskl} &18.3& 79.3& 29.8& 21.6&  52.8&  30.6&  10.5& 76.2 &18.5& 18.9 &82.7 &30.8\\
    				\textbf{DCN} \cite{calibration_generalized} & -- & -- & -- & 28.4 & 60.7 & 38.7 &14.2& 75.0 & 23.9 & 25.5 & 84.2 & 39.1\\
    				\hline \hline
    				 \textbf{CVAE-ZSL}\cite{mishra2017generative} & -- & --& 47.2 &--&-- & 34.5 &--&-- & -- & --&-- & 51.2 \\
    				\textbf{f-CLSWGAN} \cite{xian2018feature} &  61.4& 57.9& 59.6 & 43.7 & 57.7 & 49.7 &--&--&--& 57.9 & 61.4 & 59.6\\
    				\textbf{SP-AEN} \cite{Chen_2018_CVPR} &--&--&--& 34.7 &  70.6 & 46.6 &13.7&63.4&22.6& 23.3 & 90.9 & 37.1 \\
    				\textbf{cycle-UWGAN} \cite{cycle-consistancy} & -- & -- & -- & 47.9 & 59.3 & 53.0 &--&--& -- & 59.6 & 63.4 & 59.8 \\
    				\textbf{SE-GZSL} \cite{vermageneralized} & 56.3 & 67.8 & 61.5 & {41.5} & {53.3} & {46.7} & -- & --& -- & {58.3} & {68.1}  &  {62.8}\\
    				\textbf{F-VAEGAND2} \cite{fvaegan} & -- & -- & -- & 48.4  & 60.1 & 53.6 & -- & --& -- & 57.6  & 70.6 & 63.5\\
    				\textbf{VSE-S} \cite{saligram_cvpr19} & -- & -- & -- & 33.4 & 87.5& 48.4 & 24.5 & 72.0 & 36.6 & 41.6 & 91.3 &  57.2\\
    				\hline \hline 
    				\textbf{ZSML} Softmax (Ours)
    				& 57.4 & {71.1} &\textbf{63.5} &60.0 & 52.1 & \textbf{55.7}& 36.3 & 46.6 & \textbf{40.9} & 58.9 & 74.6 & \textbf{65.8} \\
    				\hline
    			\end{tabular}
    		}
    		\caption{Accuracy for GZSL, on novel proposed split (PS). U and S represent top-1 accuracy on unseen and seen class with all the $S+U$ classes. H stands for the harmonic mean. All compared methods use CNN-RNN feature for CUB dataset.}
    		\label{tab:gzsl}
    		\vspace{-13pt}
    	\end{table*}
    	\vspace{-5pt} 
    	\subsection{Zero-Shot Learning}
    	For the ZSL setting, we first train our model on seen class examples $\mathcal{D}^S$ and then synthesize samples from the unseen classes. These synthesized samples are further used to train either a multi-class linear SVM or a softmax classifier. The trained model over the synthesized examples is used to predict the classes for the test examples $\mathcal{D}^U$. We report results with both softmax classifier and linear SVM but we can, in principle, use any supervised classifier to train the model once we have generated the data. The average per-class accuracy is used as the standard evaluation metric \cite{xian2018zero}, shown in Table \ref{tab:zsl}, as it overcomes the biases towards some particular class that has more data. In the ZSL setting, our model yields 4.5\%, 6.0\%, 9.8\%, and 27.9\%  relative improvements over the current state-of-the-art on CUB, AWA1, AWA2, and aPY datasets, respectively. While, on the SUN dataset, it is very competitive as compared to the previous state-of-the-art methods. The SUN dataset contains 717 fine-grain classes; therefore, using the GAN based generation is highly prone to mode collapse. We believe that mode collapse is the possible reason for lower performance on SUN dataset. We are using the same network architecture and hyper-parameter for all the dataset. Since SUN dataset is fairly different compare to the other datasets, we believe that better hyper-parameter tuning for SUN dataset may improve the result.
    	
    	\subsection{Generalized Zero-Shot Learning}
    	Standard ZSL assumes that all test inputs are from the unseen classes. The more challenging \emph{generalized} Zero-Shot Learning (GZSL) relaxes this assumption and requires performing classification where the test set can potentially contain classes from the seen classes along with the unseen classes. We used the harmonic (HM) mean of the seen and unseen, average per class accuracy as the evaluation metric to report the results.  It is found that HM \cite{xian2018zero} is a better evaluation metric for GZSL since it overcomes the biases of predictions towards the seen class. 
    	
    	For GZSL task, we evaluate our model on the popular benchmark datasets CUB, aPY, AWA1 and AWA2. The results for GZSL is shown in Table~\ref{tab:gzsl}. Our results demonstrate that ZSML achieves significant improvements in the harmonic mean. In terms of HM based accuracies, our ZSML yields 3.9\%, 11.8\%, 3.3\% and 3.6\% relative improvement over the current state-of-the-art on CUB, aPY, AWA1 and AWA2 datasets, respectively. Thus, ZSML not only works well in the standard ZSL setting but also in the GZSL setting. From Table~\ref{tab:zsl}~and~\ref{tab:gzsl}, it is clear that all the models that show good results on the ZSL setup fail badly on the GZSL setup, whereas our model ZSML has consistently strong performance in both settings.
    	
    	\begin{table}[!t]
    		\small
    		\begin{center}
    			\scalebox{1}{
    				\addtolength{\tabcolsep}{6.0pt}
    				\begin{tabular}{|l|l|l|l|l|l|l|l|}
    					\hline
    					\multirow{2}{*}{\textbf{Method}} & \multirow{2}{*}{\textbf{N}}& \multicolumn{3}{c|}{\textbf{AwA2}} & \multicolumn{3}{c|}{\textbf{CUB}} \\ \cline{3-8} 
    					& & \textbf{U} & \textbf{S} & \textbf{H} & \textbf{U} & \textbf{S} & \textbf{H}  \\ \hline \hline 
    					\multirow{2}{*}{cycle-UWGAN \cite{cycle-consistancy}}&5 & 40.4 & 43.3 & 41.8 & 22.6 & 40.5 & 29.0 \\ &10 & 45.5 & 50.9 & 48.0 & 25.5 & 42.1  & 32.5 \\ 
    					\hline
    					\multirow{2}{*}{f-CLSWGAN  \cite{xian2018feature}} & 5& 37.8 & 44.2 & 40.7 &  30.4 & 28.5 & 29.4 \\ &10 & 40.5 & 55.9 & 46.9 & 34.7 & 38.9 & 36.6 \\ 
    					\hline
    					\multirow{2}{*}{GF-ZSL \cite{vermageneralized}} &5 & 38.2 & 44.3 & 41.0 & 29.4 & 33.0 & 31.0 \\ &10 & 41.4 & 45.1 & 43.1 & 35.6 & 43.5 & 39.1\\ 
    					\hline
    					\hline
    					\multirow{2}{*}{\textbf{Ours (ZSML)}}& 5 & {38.4} & {61.3} & {\bf 47.3}& {32.9}& {38.2}  & {\bf 35.3}\\ &10 &  {47.8} & {59.6} & {\bf 53.1}& {42.7} & {45.1} &  {\bf 43.9}\\
    					\hline
    				\end{tabular}
    			}
    			\vspace{5pt}
    			\caption{GZSL results using only five and ten example per seen classes to train the model}
    			\label{tab:abgzsl}
    			
    		\end{center}
    	\end{table}
    		\begin{figure}[!htb]
    			\vspace{-10pt}
    			\centering
    			\includegraphics[scale=0.5]{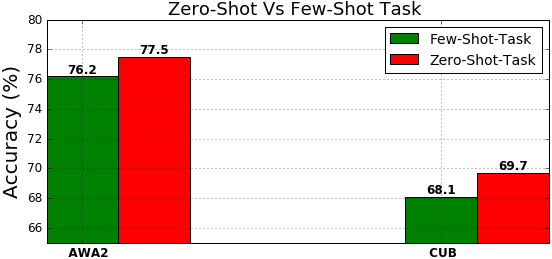}
    			\caption{Our ZSL result for AWA2 and CUB datasets with the proposed zero-shot task distribution.}
    			\label{fig:zsml_maml_split}
    			\vspace{-3pt}
    		\end{figure}
    	\subsection{Ablation Study}
    
    	In this section, we perform various ablation studies to assess the different aspects of our ZSML model on CUB, aPY and AWA2 datasets. We find that the proposed zero-shot meta-learning protocol (i.e., how we split the data from each task into meta-train and meta-validations sets) and meta-learning based adversarial generation are the key contributors for improving the model performance. We also conduct experiments when only few examples (say 5 or 10) are available from the seen class.
    	
\begin{figure}[t]
    		\includegraphics[scale=0.34]{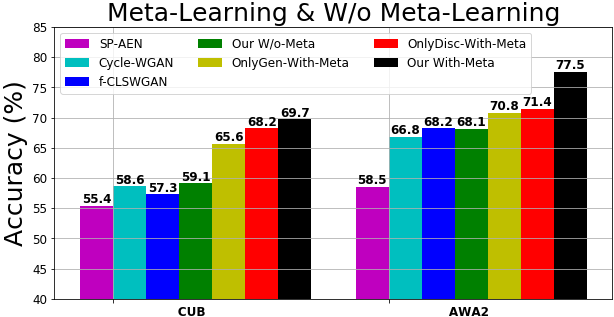} 
    		\includegraphics[scale=0.25]{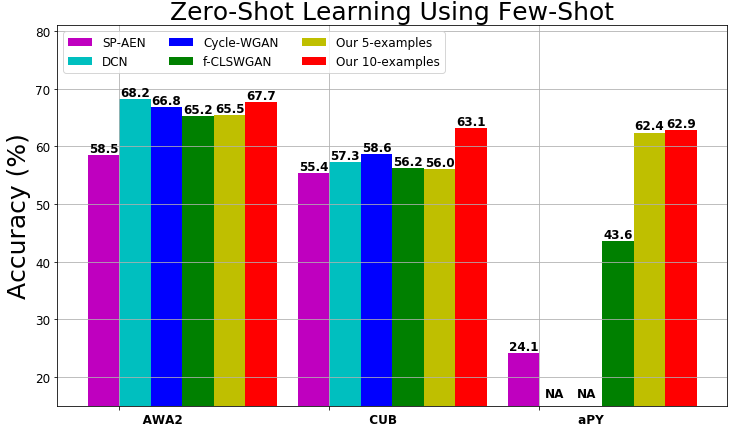}
    		\vspace{-4pt}
    		\caption{\textbf{Left:} Comparison of ZSL results on AWA2 and CUB dataset with recently proposed models based on GAN and our meta-learned GAN. \textbf{Right:} Our ZSL result when only few samples (say 5 and 10) from the seen class, while competitor uses all training samples.}
    		\vspace{-10pt}
    		\label{fig:wometalearning}
    	\end{figure}

    	\noindent \textbf{Meta-learner vs Plain-learner:} We found that meta-learning based training is the key component to boost the model performance. Meta-learned model in the adversarial setting generates high-quality samples that are close to the real samples. In Figure~\ref{fig:wometalearning}, we are comparing the results with a recent approach \cite{Chen_2018_CVPR,cycle-consistancy,xian2018feature} that uses Improved-WGAN \cite{improved-wgan} for the same problem. 
    	
    	To show the effectiveness of the proposed model, we are not using any advanced GAN architecture. We simply rely on the WGAN architecture. In the proposed model, the plain WGAN is associated with \emph{meta-learning} agents. We have found that \emph{meta-learning} framework is the key component to improve the performance. The proposed meta-learning framework improved the results in the ZSL setup, from $59.1\%$ to $69.7\%$ and $68.2\%$ to $77.5\%$ on CUB and AWA2 dataset respectively, compared to the current state-of-the-art as shown in Figure~\ref{fig:wometalearning} (Top). Also in the same setting, our approach \emph{without meta-learning} shows the ZSL results of $68.1\%$ and $59.1\%$ on AWA2 and CUB dataset respectively.

    	\noindent \textbf{Few-Shot ZSL and Few-Shot GZSL:} This is another significant result of the proposed approach. The meta-learning framework is specially designed for few-shot learning. So it is natural to ask how ZSL/GZSL will perform when only few-shot are present from the seen classes. This is the \emph{most extreme} case for any classification algorithm (i.e. only a few examples are present from the seen class and at test time we have unseen/novel data). We perform the experiment for AWA2, CUB and aPY datasets assuming that only 5 or 10 examples per seen class are available and unseen class has no data at training time. In the 5 examples per class experiment, we create a new dataset (by sampling from the original dataset) that contains 5 examples per seen classes (i.e. for 40 unseen classes in AWA2 dataset, our new dataset contains only $5\times40=200$ samples). The model learns to generate unseen samples when it sees only 5 examples per seen class. Once the model is trained, we perform the classification following the procedure mentioned in Subsection~\ref{sec:gen_classification}. We follow the same process for 10 examples per seen class.

    	As shown in Figure~\ref{fig:wometalearning} (Bottom), with as few as only 10 examples per-class our approach outperforms other state-of-the-art methods on CUB, aPY and AWA2 datasets in ZSL setting, also using only 5 examples per class our result are very competitive (while competitor model uses \emph{all examples} in training). Also as shown in Table~\ref{tab:abgzsl}, in the most challenging GZSL setting, using only 5 or 10 samples our result out performs the recent approach by a significant margin.
    	
    	\noindent \textbf{Zero-Shot MAML Split vs Traditional MAML Split:}
    	We propose a novel task distribution for ZSML where each task $\Ti$ is partitioned into two sets $\Ttr$ and $\Tval$ and the classes in $\Ttr$ and $\Tval$ are disjoint. While in the MAML setup these classes are the same. This disjoint class partition helps in \emph{learning to generate the novel classes in the training itself}. The ablation over the MAML and ZSML task distribution is shown in Figure~\ref{fig:zsml_maml_split}. The proposed training set and validation set split (per episode) performs significantly better than traditional MAML split. Using the novel ZSML split, the ZSL results improves 1.7\% and 2.4\% on the AWA2 and CUB dataset, respectively.
    	
    	\noindent \textbf{Which Aspects Benefit More from Meta-Adversarial Learning?} In adversarial learning, the sample quality depends on how powerful the discriminator and generator are.
    	The optimal discriminator minimizes the JS-Divergence between the generated and the original samples~\cite{GAN}. The meta-learner associated with discriminator or generator provides a powerful discriminator and generator by enhancing their learning capability. The optimal discriminator provides strong feedback to the generator and the generator continuously increases its generation capability. We observe that if we remove the meta-learner from the discriminator, we have 5.8\% and 8.6\% accuracy drop as compared to our model with a meta-learning component on CUB and AWA2 dataset, respectively. The significant accuracy drop occurs since the discriminator is not optimal and provides poor feedback to the generator. Even though we have a much more powerful generator, because of the poor feedback from the weak discriminator, the generator is unable to learn. Similarly, if we remove the meta-learner from the generator, we again observe a significant accuracy drop (2.2\% and 7.9\% on CUB and AWA2 dataset, respectively). Since the generator has a reduced capability without meta-learner, even though discriminator provides strong feedback to the generator, the generator is not powerful enough to counter the discriminator. Also, if we remove the meta-learning agent from generator and discriminator, it becomes a plain adversarial network. The ablation results are shown in Figure~\ref{fig:wometalearning}. More ablation are provided into supplementary material. 
    	\vspace{-8pt}
    	\section{Conclusion}
    	In this work, we identify and address three key limitations of current ZSL approaches, that limit the performance of the recent generative models for ZSL/GZSL. We observe that a meta-learning based approach can naturally overcome these limitations in a principled manner. We have proposed a novel framework for ZSL and GZSL which is based on the meta-learning framework over a conditional generative model (WGAN). We also propose a novel zero-shot task distribution for the meta-learning model to mimic the ZSL behaviour. We have conducted extensive experiments benchmark ZSL datasets. In the few-shot, as well as standard GZSL setting, the proposed model outperforms the state-of-the-art methods by a significant margin. Our ablation study shows that the proposed meta-learning framework and zero-shot task distribution are the key components for performance improvement. Finally, although our focus here has been on ZSL and generalized ZSL, our meta-learning based adversarial generation model can be useful for the problem of distribution learning and generation tasks as well~\cite{reed2017few,hewitt2018variational}. For GZSL, we achieve the state-of-the-art results over all the standard datasets, whereas for ZSL, we surpass the state-of-art results by a significant margin on aPY, CUB, AWA1 and AWA2 datasets.  
    	\newpage
    	
    	\small
    	\bibliographystyle{abbrv}
    	\bibliography{sample.bib}

\begin{thebibliography}{}

\bibitem[Akata et~al., 2013]{akata2013label}
Akata, Z., Perronnin, F., Harchaoui, Z., and Schmid, C. (2013).
\newblock Label-embedding for attribute-based classification.
\newblock In {\em CVPR}, page 819.

\bibitem[Akata et~al., 2015]{SJE}
Akata, Z., Reed, S., Walter, D., Lee, H., and Schiele (2015).
\newblock Evaluation of output embeddings for fine-grained image
  classification.
\newblock In {\em CVPR}, pages 2927--2936.

\bibitem[Arjovsky et~al., 2017]{wgan}
Arjovsky, M., Chintala, S., and Bottou, L. (2017).
\newblock Wasserstein gan.
\newblock {\em arXiv preprint arXiv:1701.07875}.

\bibitem[Bucher et~al., 2017]{bucher2017generating}
Bucher, M., Herbin, S., and Jurie, F. (2017).
\newblock Generating visual representations for zero-shot classification.
\newblock In {\em ICCV, Workshop}.

\bibitem[Changpinyo et~al., 2016]{changpinyo2016synthesized}
Changpinyo, S., Chao, W.-L., Gong, B., and Sha, F. (2016).
\newblock Synthesized classifiers for zero-shot learning.
\newblock In {\em CVPR}, pages 5327--5336.

\bibitem[Chao et~al., 2016]{chao2016empirical}
Chao, W.-L., Changpinyo, S., Gong, B., and Sha, F. (2016).
\newblock An empirical study and analysis of generalized zero-shot learning for
  object recognition in the wild.
\newblock In {\em ECCV}.

\bibitem[Chen et~al., 2018]{Chen_2018_CVPR}
Chen, L., Zhang, H., Xiao, J., Liu, W., and Chang, S.-F. (2018).
\newblock Zero-shot visual recognition using semantics-preserving adversarial
  embedding networks.
\newblock In {\em CVPR}.

\bibitem[Clou{\^a}tre and Demers, 2019]{few-shotimage}
Clou{\^a}tre, L. and Demers, M. (2019).
\newblock Figr: Few-shot image generation with reptile.
\newblock {\em arXiv preprint arXiv:1901.02199}.

\bibitem[Farhadi et~al., 2009]{farhadi2009describing}
Farhadi, A., Endres, I., Hoiem, D., and Forsyth, D. (2009).
\newblock Describing objects by their attributes.
\newblock In {\em CVPR}, pages 1778--1785. IEEE.

\bibitem[Felix et~al., 2018]{cycle-consistancy}
Felix, R., Vijay~Kumar, B., Reid, I., and Carneiro, G. (2018).
\newblock Multi-modal cycle-consistent generalized zero-shot learning.
\newblock {\em ECCV}.

\bibitem[Finn et~al., 2017]{finn2017model}
Finn, C., Abbeel, P., and Levine, S. (2017).
\newblock Model-agnostic meta-learning for fast adaptation of deep networks.
\newblock {\em ICML}.

\bibitem[Frome et~al., 2013]{frome2013devise}
Frome, A., Corrado, G.~S., Shlens, J., Bengio, S., Dean, J., Mikolov, T.,
  et~al. (2013).
\newblock Devise: A deep visual-semantic embedding model.
\newblock In {\em NIPS}, pages 2121--2129.

\bibitem[Gan et~al., 2015]{gan2015exploring}
Gan, C., Lin, M., Yang, Y., Zhuang, Y., and Hauptmann, A.~G. (2015).
\newblock Exploring semantic inter-class relationships (sir) for zero-shot
  action recognition.
\newblock In {\em Twenty-ninth AAAI conference on artificial intelligence}.

\bibitem[Goodfellow et~al., 2014]{GAN}
Goodfellow, I., Pouget-Abadie, J., Mirza, M., Xu, B., Warde-Farley, D., Ozair,
  S., Courville, A., and Bengio, Y. (2014).
\newblock Generative adversarial nets.
\newblock In {\em NIPS}, pages 2672--2680.

\bibitem[Gulrajani et~al., 2017]{improved-wgan}
Gulrajani, I., Ahmed, F., Arjovsky, M., Dumoulin, V., and Courville, A.~C.
  (2017).
\newblock Improved training of wasserstein gans.
\newblock In {\em NIPS}, pages 5767--5777.

\bibitem[Guo et~al., 2017]{guo2017synthesizing}
Guo, Y., Ding, G., Han, J., and Gao, Y. (2017).
\newblock Synthesizing samples for zero-shot learning.
\newblock In {\em IJCAI}.

\bibitem[Hewitt et~al., 2018]{hewitt2018variational}
Hewitt, L.~B., Nye, M.~I., Gane, A., Jaakkola, T., and Tenenbaum, J.~B. (2018).
\newblock The variational homoencoder: Learning to learn high capacity
  generative models from few examples.
\newblock {\em arXiv preprint arXiv:1807.08919}.

\bibitem[Hu et~al., 2018]{hu2018correction}
Hu, R.~L., Xiong, C., and Socher, R. (2018).
\newblock Correction networks: Meta-learning for zero-shot learning.

\bibitem[Karras et~al., 2018]{progressivekarras}
Karras, T., Aila, T., Laine, S., and L, J. (2018).
\newblock Progressive growing of gans for improved quality, stability, and
  variation.
\newblock {\em ICLR}.

\bibitem[Khare et~al., 2019]{khare2019generative}
Khare, V., Mahajan, D., Bharadhwaj, H., Verma, V., and Rai, P. (2019).
\newblock A generative framework for zero-shot learning with adversarial domain
  adaptation.
\newblock {\em arXiv preprint arXiv:1906.03038}.

\bibitem[Kingma and Welling, 2014]{VAE}
Kingma, D.~P. and Welling, M. (2014).
\newblock Auto-encoding variational bayes.
\newblock {\em ICLR}.

\bibitem[Kodirov et~al., 2017]{SAE2017}
Kodirov, E., Xiang, T., and Gong, S. (2017).
\newblock Semantic autoencoder for zero-shot learning.
\newblock {\em CVPR}.

\bibitem[Kumar~Verma et~al., 2019]{kumar2019generative}
Kumar~Verma, V., Mishra, A., Mishra, A., and Rai, P. (2019).
\newblock Generative model for zero-shot sketch-based image retrieval.
\newblock In {\em Proceedings of the IEEE Conference on Computer Vision and
  Pattern Recognition Workshop}, pages 0--0.

\bibitem[Lampert et~al., 2009]{lampert2009learning}
Lampert, C.~H., Nickisch, H., and Harmeling, S. (2009).
\newblock Learning to detect unseen object classes by between-class attribute
  transfer.
\newblock In {\em CVPR}, pages 951--958. IEEE.

\bibitem[Lampert et~al., 2014]{IAP}
Lampert, C.~H., Nickisch, H., and Harmeling, S. (2014).
\newblock Attribute-based classification for zero-shot visual object
  categorization.
\newblock {\em PAMI}, 36(3):453--465.

\bibitem[Li et~al., 2019]{lisgan}
Li, J., Jin, M., Lu, K., Ding, Z., Zhu, L., and Huang, Z. (2019).
\newblock Leveraging the invariant side of generative zero-shot learning.
\newblock {\em CVPR}.

\bibitem[Liu et~al., 2018]{calibration_generalized}
Liu, S., Long, M., Wang, and Jordan, M.~I. (2018).
\newblock Generalized zero shot learning with deep calibration network.
\newblock In {\em NIPS}, pages 2006--16.

\bibitem[Mandal et~al., 2019]{mandal2019out}
Mandal, D., Narayan, S., Dwivedi, S.~K., Gupta, V., Ahmed, S., Khan, F.~S., and
  Shao, L. (2019).
\newblock Out-of-distribution detection for generalized zero-shot action
  recognition.
\newblock In {\em Proceedings of the IEEE Conference on Computer Vision and
  Pattern Recognition}, pages 9985--9993.

\bibitem[Mishra et~al., 2017]{mishra2017generative}
Mishra, A., Reddy, M., Mittal, A., and Murthy, H.~A. (2017).
\newblock A generative model for zero shot learning using conditional
  variational autoencoders.
\newblock {\em CVPR Workshop}.

\bibitem[Mishra et~al., 2018]{mishra2018generative}
Mishra, A., Verma, V.~K., Reddy, M., Rai, P., and Mittal, A. (2018).
\newblock A generative approach to zero-shot and few-shot action recognition.
\newblock {\em WACV-18, pp. 372-380.}

\bibitem[Nayak and Chakraborty, 2019]{nayak2019zero}
Nayak, G.~K., M. K. R. S. V. B. R.~V. and Chakraborty, A. (2019).
\newblock Zero-shot knowledge distillation in deep networks.
\newblock In {\em International Conference on Machine Learning}, pages
  4743--4751.

\bibitem[Norouzi et~al., 2013]{conse}
Norouzi, M., Mikolov, T., Bengio, S., Singer, Y., Shlens, J., Frome, A.,
  Corrado, G.~S., and Dean, J. (2013).
\newblock Zero-shot learning by convex combination of semantic embeddings.
\newblock {\em NIPS}.

\bibitem[Pal and Balasubramanian, 2019a]{Zero-ShotTaskTransfer}
Pal, A. and Balasubramanian, V.~N. (2019a).
\newblock Zero-shot task transfer.
\newblock In {\em The IEEE Conference on Computer Vision and Pattern
  Recognition (CVPR)}.

\bibitem[Pal and Balasubramanian, 2019b]{zero-shottask_cvpr19}
Pal, A. and Balasubramanian, V.~N. (2019b).
\newblock Zero-shot task transfer.
\newblock {\em CVPR}.

\bibitem[Ravi and Larochelle, 2016]{ravi2016optimization}
Ravi, S. and Larochelle, H. (2016).
\newblock Optimization as a model for few-shot learning.
\newblock {\em ICLR}.

\bibitem[Reed et~al., 2016]{reed2016learning}
Reed, S., Akata, Z., Lee, H., and S, B. (2016).
\newblock Learning deep representations of fine-grained visual descriptions.
\newblock In {\em CVPR}, pages 49--58.

\bibitem[Reed et~al., 2017]{reed2017few}
Reed, S., Chen, Y., Paine, T., Oord, A. v.~d., Eslami, S., Rezende, D.,
  Vinyals, O., and de~Freitas, N. (2017).
\newblock Few-shot autoregressive density estimation: Towards learning to learn
  distributions.
\newblock {\em ICLR}.

\bibitem[Romera and Torr, 2015a]{ESZSL2015}
Romera, Paredes, B. and Torr, P. (2015a).
\newblock An embarrassingly simple approach to zero-shot learning.
\newblock In {\em ICML}, pages 2152--2161.

\bibitem[Romera and Torr, 2015b]{romera2015embarrassingly}
Romera, Paredes, B. and Torr, P.~H. (2015b).
\newblock An embarrassingly simple approach to zero-shot learning.
\newblock In {\em ICML}, pages 2152--2161.

\bibitem[Russakovsky et~al., 2015]{imagenet2015}
Russakovsky, O., Deng, J., Su, H., Krause, J., Satheesh, S., Ma, S., Huang, Z.,
  Karpathy, A., Khosla, A., Bernstein, M., et~al. (2015).
\newblock Imagenet large scale visual recognition challenge.
\newblock {\em IJCV}, pages 211--252.

\bibitem[Schonfeld et~al., 2019]{cada-vae}
Schonfeld, E., Ebrahimi, S., Sinha, S., Darrell, T., and Akata, Z. (2019).
\newblock Generalized zero-and few-shot learning via aligned variational
  autoencoders.
\newblock In {\em CVPR}, pages 8247--8255.

\bibitem[Shen et~al., 2018]{shen2018zero}
Shen, Y., Liu, L., Shen, F., and Shao, L. (2018).
\newblock Zero-shot sketch-image hashing.
\newblock In {\em Proceedings of the IEEE Conference on Computer Vision and
  Pattern Recognition}, pages 3598--3607.

\bibitem[Snell et~al., 2017]{prototypical}
Snell, J., Swersky, K., and Zemel, R. (2017).
\newblock Prototypical networks for few-shot learning.
\newblock In {\em NIPS}, pages 4077--4087.

\bibitem[Socher et~al., 2013]{cmt}
Socher, R., Ganjoo, M., Manning, C.~D., and Ng, A. (2013).
\newblock Zero-shot learning through cross-modal transfer.
\newblock In {\em NIPS}, pages 935--943.

\bibitem[Song et~al., 2018]{song2018transductive}
Song, J., Shen, C., Yang, Y., Liu, Y., and S, M. (2018).
\newblock Transductive unbiased embedding for zero-shot learning.
\newblock In {\em CVPR}, pages 1024--1033.

\bibitem[Verma et~al., 2018]{vermageneralized}
Verma, V.~K., Arora, G., Mishra, A., and Rai, P. (2018).
\newblock Generalized zero-shot learning via synthesized examples.
\newblock {\em CVPR}.

\bibitem[Verma and Rai, 2017]{verma2017simple}
Verma, V.~K. and Rai, P. (2017).
\newblock A simple exponential family framework for zero-shot learning.
\newblock In {\em ECML-PKDD}, pages 792--808. Springer.

\bibitem[Vinyals et~al., 2016]{matching}
Vinyals, O., Blundell, C., Lillicrap, T., Wierstra, D., et~al. (2016).
\newblock Matching networks for one shot learning.
\newblock In {\em NIPS}, pages 3630--3638.

\bibitem[Wang et~al., 2018]{wang2017zero}
Wang, W., Pu, Y., Verma, V.~K., Fan, K., Zhang, Y., Chen, C., Rai, P., and
  Carin, L. (2018).
\newblock Zero-shot learning via class-conditioned deep generative models.
\newblock {\em AAAI}.

\bibitem[Welinder et~al., 2010]{welinder2010caltech}
Welinder, P., Branson, S., Mita, T., Wah, C., Schroff, F., Belongie, S., and
  Perona, P. (2010).
\newblock Caltech-ucsd birds 200.
\newblock {\em California Institute of Technology}.

\bibitem[Xian et~al., 2016]{latem}
Xian, Y., Akata, Z., Sharma, G., Nguyen, Q., Hein, M., and Schiele, B. (2016).
\newblock Latent embeddings for zero-shot classification.
\newblock In {\em CVPR}, pages 69--77.

\bibitem[Xian et~al., 2018a]{xian2018zero}
Xian, Y., Lampert, C.~H., Schiele, B., and Akata, Z. (2018a).
\newblock Zero-shot learning-a comprehensive evaluation of the good, the bad
  and the ugly.
\newblock {\em PAMI}.

\bibitem[Xian et~al., 2018b]{xian2018feature}
Xian, Y., Lorenz, T., Schiele, B., and Akata, Z. (2018b).
\newblock Feature generating networks for zero-shot learning.
\newblock In {\em CVPR}.

\bibitem[Xian et~al., 2019]{fvaegan}
Xian, Y., Sharma, S., Schiele, B., and Akata, Z. (2019).
\newblock f-vaegan-d2: A feature generating framework for any-shot learning.
\newblock In {\em CVPR}, pages 10275--10284.

\bibitem[Xiao et~al., 2010]{xiao2010sun}
Xiao, J., Hays, J., Ehinger, K.~A., Oliva, A., and Torralba, A. (2010).
\newblock Sun database: Large-scale scene recognition from abbey to zoo.
\newblock In {\em CVPR, 2010}, pages 3485--3492.

\bibitem[Xu et~al., 2015]{xu2015semantic}
Xu, X., Hospedales, T., and Gong, S. (2015).
\newblock Semantic embedding space for zero-shot action recognition.
\newblock In {\em 2015 IEEE International Conference on Image Processing
  (ICIP)}, pages 63--67. IEEE.

\bibitem[Xu et~al., 2017]{xu2017transductive}
Xu, X., Hospedales, T., and Gong, S. (2017).
\newblock Transductive zero-shot action recognition by word-vector embedding.
\newblock {\em International Journal of Computer Vision}, pages 1--25.

\bibitem[Zhang and Koniusz, 2018]{zskl}
Zhang, H. and Koniusz, P. (2018).
\newblock Zero-shot kernel learning.
\newblock In {\em CVPR}, pages 7670--7679.

\bibitem[Zhang et~al., 2017]{dem}
Zhang, L., Xiang, T., and Gong, S. (2017).
\newblock Learning a deep embedding model for zero-shot learning.
\newblock In {\em CVPR}, pages 3010--3019.

\bibitem[Zhang and Saligrama, 2015]{sse}
Zhang, Z. and Saligrama, V. (2015).
\newblock Zero-shot learning via semantic similarity embedding.
\newblock In {\em ICCV}, pages 4166--4174.

\bibitem[Zhang and Saligrama, 2016]{saligram2016learningJoint}
Zhang, Z. and Saligrama, V. (2016).
\newblock Learning joint feature adaptation for zero-shot recognition.
\newblock {\em arXiv preprint arXiv:1611.07593}.

\bibitem[Zhu et~al., 2019]{saligram_cvpr19}
Zhu, Pengkai, Wang, H., and Saligrama, V. (2019).
\newblock Generalized zero-shot recognition based on visually semantic
  embedding.
\newblock In {\em CVPR}, pages 2995--3003.

\bibitem[Zhu et~al., 2018]{noisy_text}
Zhu, Y., Elhoseiny, M., Liu, B., Peng, X., and Elgammal, A. (2018).
\newblock A generative adversarial approach for zero-shot learning from noisy
  texts.
\newblock In {\em CVPR}, pages 1004--1013.

\end{thebibliography}
    	
    	\newpage
    	\section*{Appendix}
    	\appendix

    		\section{Datasets}
    		This section describes the benchmark datasets used for model evaluation over ZSL and GZSL setup. We evaluate our proposed method on five benchmark datasets. SUN and CUB are fine-grained datasets, and each class has limited data that makes this dataset very challenging. AWA1 and AWA2 are animal datasets with a diverse background. aPY is a small scale dataset, but the diverse domain in seen and unseen class makes the dataset very challenging. The objective of the proposed approach is not to generate the seen/unseen image but the ResNet-101 feature vector. The objective of the proposed approach is to produce state-of-the-art result for ZSL and GZSL setting. Therefore, like the other recent models \cite{vermageneralized,xian2018feature,xian2018zero,cycle-consistancy}, our objective is to synthesize high-quality image features. We are using ResNet-101 feature vectors for the class attributes as used by the other competitive approaches. ResNet-101 model is pretrained on the ImageNet \cite{imagenet2015} dataset. The features for all the datasets are extracted using the pretrained  ResNet-101 model without any further finetuning. Also, the seen and unseen class split is done such that no test/unseen classes are present in the ImageNet dataset; otherwise, it violates the ZSL setting \cite{xian2018zero}. The complete dataset with the train, validation, and test splits are provided by \cite{xian2018zero}. We used the same setup as used by other approaches (mentioned in Table-1 and Table-3 in the main paper). Table-\ref{tab:data} below summarizes the statistics of all the datasets.
    		
    		\begin{table}[htb]
    			\small
    			\centering
    			\scalebox{1}{
    				\addtolength{\tabcolsep}{21pt}
    				\begin{tabular}{||l|c|c|c||} 
    					\hline
    					Dataset & Attribute/Dim & \#Image & Seen/Unseen Class \\ [0.5ex] 
    					\hline\hline
    					AWA1 & A/85 & 30475 & 40/10 \\
    					AWA2 & A/85 & 37322 & 40/10 \\
    					CUB & CR/1024 & 11788 & 150/50 \\
    					SUN & A/102 & 14340 & 645/72 \\
    					aPY & A/64 & 15339 & 20/12 \\
    					\hline
    					\hline
    				\end{tabular}
    			}
    			\vspace{5pt}
    			\caption{Datasets used in our experiments and their statistics. CR: CNN-RNN \cite{reed2016learning}} \label{tab:data}
    		\end{table}
    		
    		\subsection{Animals with Attributes (AWA)}
    		In AWA1 dataset \cite{lampert2009learning}, there are 30,475 images in total. There are 50 classes of animals captured in a diverse background making the dataset very challenging. In the ZSL setting, 40 classes are used for training and validation, and the rest of the 10 classes are used for testing. The dataset also contains an 85-dimensional attribute vector provided by a human annotator. There are two types of attribute vector with the AWA dataset, binary and continuous. The continuous attributes are much informative as used by other models. The raw images of the AWA1 dataset are not provided, and  only the features are available. Therefore another updated version, AWA2, is released with the raw images as well. In our experiment, we evaluate the model using both the datasets and perform the ablation over the AWA2 dataset. The ResNet-101 feature is used for both the datasets pretrained on ImageNet dataset. Similar to other approaches, no fine-tuning is performed for the seen classes, and the split is done in such a way that no unseen class belongs to the ImageNet classes.
    		
    		\subsection{Caltech UCSD Birds 200 (CUB)}
    		The CUB \cite{welinder2010caltech} dataset comprises of 11,788 images of birds in total which belongs to 200 classes. In the ZSL setting, 150 classes are used for training and validation, while 50 classes are used for testing. The CUB dataset is a fine-grained dataset containing 200 classes of birds, and each class has nearly 60 samples. Some of the classes are very similar even for humans, making it is very challenging to detect the birds correctly. To collect large samples from each class is very challenging. Therefore, each class contains a limited sample. For deep learning algorithms, it is a difficult task to train a model using only 60 samples per class. In this case, meta-learning models have an advantage and show significant improvement. In the CUB dataset, each class is also provided with a 312-dimensional human annotated class attribute vector. Also \cite{reed2016learning} provides the textual description for each image. Using the character-based CNN-RNN, \cite{reed2016learning} provides 1024-dimensional embedding of the textual description. Recent works use the CNN-RNN feature as the attributes since it gives superior performance compared to 312-dimensional attribute vector. Without any fine-tuning on the CUB dataset, ResNet-101 pre-trained model trained on the ImageNet is used for the feature extraction of the seen and unseen classes.
    		
    		\subsection{SUN Scene Recognition (SUN)} The SUN dataset \cite{xiao2010sun} consists of 717 scenes or classes. We used the same split proposed by \cite{xian2018zero}, where 645 classes are used for train and validation, and rest 72 unseen classes are used for testing. The split is done in such a way that no test class is from the ImageNet classes. This dataset contains 14,340 fine-grained images where each image is also associated with a human annotated attribute vector. The attribute of the same class is averaged and used the class attribute. The attribute is of 102-dimensions. Again ResNet-101 pretrained feature is used as the image feature without any fine-tuning.
    		
    		\subsection{a-Pascal a-Yahoo (aPY)} In aPY dataset \cite{farhadi2009describing}, there are 15339 total images belonging to 32 classes. The training and validation dataset contains 20 classes, and for unseen/test class 12 classes are used \cite{xian2018zero}. In aPY, each class is associated with a 64-dimensional human labeled attribute vector. Unlike the other datasets, this dataset contains very diverse objects. Therefore, for ZSL, this is a very challenging dataset. Same ResNet-101 feature is used without any fine-tuning on the aPY dataset.
    		
    		In the next section, we describe the model architecture and the experimental setup for ZSL and GZSL. We denote the part of the examples with seen classes by $\mathcal{X}^S$ and that of unseen classes by $\mathcal{X}^U$.
    		
    		\section{Model Architecture Details}
    		\label{model_details}
    		The proposed model is composed of a Generator $G$, a Discriminator $D$ and a Classifier $C$ network, also each component associated with the meta-learning agent. The meta-learner optimizes its parameters based on $\Ttr \in \Ti$. Once the inner loop is optimized, the loss over optimal parameters of the meta-learner is calculated for $\Tval \in \Ti$ data. Note that the class of $\Ttr$ and $\Tval$ are disjoint. Therefore the outer loop is optimized over the novel class. Therefore the model learns to optimize the loss over the novel class data on the outer loop. 
    		
    		The setting for each task is $N$-way $K$-shot. For all datasets, $\Ttr$ used in the inner loop is in $10$-way $5$-shot setting, but, to calculate the loss in the outer loop, $\Tval$ is in $10$-way $3$-shot setting. At test time, we have $M$-way $0$-shot meta-learner model for unseen class classification, where $M$ is the number of classes in the test examples. For ZSL, $M$ contains $U$ classes, and for GZSL, $M$ is $S+U$ classes. We have sampled 10 tasks for each batch to train the model. The learning rate in the algorithm \ref{alg:maml} uses $\eta_1=\eta_2=0.001$ and $\beta_1=\beta_2=0.00001$. For CUB dataset, we trained the model for 5000 iterations while for the aPY dataset, 500 iterations are sufficient, and the performance saturated. The AWA1 and AWA2 datasets took 20000 iterations for convergence. The architecture details for all the components of the model are as follows:
    		
    		The complete architecture is: $[Input, 2048, 2048, Output]$. The non-linearity is used after the input layer and before the output layer, and a dropout probability of 0.5 is used on all the layers. The network $D$ also contains two hidden layers with the same non-linearity as that of $G$, but no BatchNorm is used. The complete architecture is given as: $[$\textit{Input}, $1024$, $1024$, $512$, $1]$. The non-linearity is used on all the layers. The classification network $C$ contains a single layer hidden network with the same non-linearity as the previous one. The classification architecture is given as; $[Input, 512, 512, Output]$ and no BatchNorm is used.
    		
    		\subsection{Generator (G)}
    		The network $G$ contains two hidden layers of size $2048$ with BatchNorm applied on each hidden layer. For non-linearity, we use Leaky-ReLU with parameter 0.2. The output layer size is of $2048$-dimension (size of ResNet-101 feature). Dropout with probability 0.5 is used for all the layers. The details are given below:
    		
    		$[Input \rightarrow 2048 \rightarrow{Dropout (0.5)} \rightarrow LeReLU(0.2) \rightarrow 2048 \rightarrow BatchNorm \rightarrow Dropout(0.5) \rightarrow 2048 \rightarrow BatchNorm \rightarrow Dropout(0.5) \rightarrow LeReLU(0.2) \rightarrow Output (2048)]$
    		
    		\subsection{Discriminator (D)} 
    		The network $D$ contains two hidden layers with same non-linearity as that of $G$, but no batch-norm is used. The non-linearity is used on all the layers. The details are given below:
    		
    		$[Input \rightarrow 1024 \rightarrow{Dropout (0.5)} \rightarrow LeReLU(0.2) \rightarrow 1024 \rightarrow  Dropout(0.5) \rightarrow 512 \rightarrow Dropout(0.5) \rightarrow LeReLU(0.2) \rightarrow 1]$
    		
    		\subsection{Classifier (C)}
    		The classifier network $C$ contains a single layer hidden network with a Leaky-ReLU non-linearity with parameter 0.2. Dropout with probability value 0.5 is used on each layer. The details are given below:
    		
    		$[Input \rightarrow 512 \rightarrow{Dropout (0.5)} \rightarrow LeReLU(0.2) \rightarrow 512 \rightarrow Dropout(0.5) \rightarrow LeReLU(0.2) \rightarrow Output]$
    		\section{Sample generation and Classification}
    		Once the model is trained, we are generating samples from unseen classes using the class attribute. The samples are generated using the generator network conditioned on the class attributes, such that the input is a concatenation of the class attribute vector with a noise vector $\mathbf{z}\sim \mathcal{N}(0,0.25)$. While training, we used $\mathbf{z}\sim \mathcal{N}(0,0.5)$, and we empirically found that $\mathbf{z}$ with 0.25 standard deviation gives the stable result. We claim that once the samples are synthesized, we can use any supervised classifier. Therefore, to support our claim, we are reporting the results using the two most popular classifiers. We are generating 200 samples for AWA1, aPY, and AWA2 dataset while for CUB and SUN dataset 100 samples are sufficient for the stable results. Training is done using the synthesized samples, and unseen class samples are tested over the trained model.
    		\begin{table*}[t]
    			\centering
    			\scalebox{1}{
    				\addtolength{\tabcolsep}{11pt}
    				
    				\begin{tabular}{|l|  c | c | c |c|} 
    					\hline
    					\textbf{Method}&  {\textbf{CUB}} & {\textbf{AWA1}} & {\textbf{AWA2}}  & \textbf{aPY}\\ \hline
    					\hline
    					
    					\textbf{DCN} \cite{calibration_generalized}  & 56.2 & --& 65.2 & 43.6\\ 
    					\textbf{ZSKL} \cite{zskl}  & 51.7 & \textbf{70.1} & \textbf{70.5} & 45.3\\
    					\textbf{GFZSL}\cite{verma2017simple}     & 49.2  & 69.4 & 67.0 & 38.4 \\
    					\textbf{cycle-UWGAN} \cite{cycle-consistancy}  & 58.6 &-- & 66.8  &--\\
    					\textbf{f-CLSWGAN} \cite{xian2018feature}  & 57.3 & -- & 68.2 &--\\
    					\textbf{SE-ZSL} \cite{vermageneralized}  &{59.6} & {69.5} &  {69.2}&--\\
    					\hline\hline
    					\textbf{ZSML} (Ours)  5-Example per class & {56.0} & {65.1} & {65.5}  & \textbf{62.4}\\
    					\textbf{ZSML} (Ours) 10-Example per class & \textbf{63.1} & {66.3} & {67.7}  & \textbf{62.9}\\
    					\hline\hline
    					\textbf{ZSML} Softmax (Ours) All-examples & \textbf{69.6} & \textbf{73.5} & \textbf{76.1}  & \textbf{64.1}\\
    					\textbf{ZSML} SVM (Ours) All-examples & \textbf{69.7} & \textbf{74.3} & \textbf{77.5}  & \textbf{64.0}\\
    					\hline
    				\end{tabular}
    			}
    			\vspace{0.2cm}
    			\caption{Zero-Shot Learning results on the novel setup proposed by \cite{xian2018zero}. The non-generative models models are mentioned at the top and the generative models are mentioned at the bottom. All the results are in the Inductive setting.}
    			\label{tab:zsl}
    		\end{table*}
    		\subsection{SoftMax Classifier2 (C2)}
    		The classifier contains a single layer neural network without any nonlinearity. We use dropout with probability 0.5, and the output layer contains softmax. The number of classes on the output layer is the number of unseen class $U$ for ZSL, whereas it is the number of seen and unseen classes $S+U$ for GZSL. We provide the model details  below:  
    		
    		$[Input \rightarrow 2048 \rightarrow{Dropout (0.5)} \rightarrow Output]$
    		
    		\subsection{Linear-SVM}
    		We also use Linear-SVM for classification of the unseen class data. The model is trained over the synthesized samples. The samples used in training are mentioned above. Linear-SVM consistently performs better than softmax, but training Linear-SVM is very time-consuming for a larger number of classes and data size. Therefore, for GZSL, we are reporting the results over softmax only. In Linear-SVM, we used soft-margin penalty $C=1$, and class weights are balanced based on the class frequencies in the data. 
    		
    		\section{Algorithm}
    		The algorithm for the complete approach is given below\footnote{We will provide code and data upon publication}:
    		\begin{algorithm}[!htb]
    			\caption{Generative Adversarial MAML for ZSL}
    			\label{alg:maml}
    			\begin{algorithmic}[1]
    				\Require $p(\mathcal{T})$: distribution over tasks
    				\Require $\eta_1,\eta_2$, $\beta_1,\beta_2$: step size hyperparameters
    				\State randomly initialize $\theta_d$ and $\theta_{gc}$
    				\While{not done}
    				\State Sample batch of tasks $\Ti \sim p(\mathcal{T}): \Ti=\{\Ttr,\Tval\}$ with \underline{\textbf{disjoint}} set of -
    				\State classes between $\Ttr,\Tval$ are used
    				\ForAll{$\Ti$}
    				\State Evaluate $\nabla_{\theta_d} l_{\Ti}^D(\theta_{d})$ with respect to $\Ttr \in \Ti$
    				\State Evaluate $\nabla_{\theta_{gc}} l_{\Ti}^{GC}(\theta_{gc})$ with respect to $\Ttr \in \Ti$
    				\State Compute adapted parameters: $\theta_d'=\theta_d+ \eta_1 \nabla_{\theta_d} l_{\Ti}^D(\theta_{d})$
    				\State Compute adapted parameters: $\theta_{gc}'=\theta_{gc}- \eta_2 \nabla_{\theta_{gc}} l_{\Ti}^{GC}(\theta_{gc})$
    				\EndFor
    				\State Update $\theta_d=\theta_d+ \beta_1 \nabla_{\theta_d} \sum_{ \Tval \in \Ti}l^D_{\Tval}(\theta_{d}')$
    				\State Update $\theta_{gc}=\theta_{gc}- \beta_2 \nabla_{\theta_{gc}} \sum_{ \Tval \in \Ti } l^{GC}_{\Tval}(\theta_{gc}')$
    				\EndWhile
    			\end{algorithmic}
    		\end{algorithm}
    		
    		\section{Ablation}
    		In this section, we are performing the ablation study with a different setup. The CUB-200 and AWA2 datasets are used for the ablation analysis over different components. In ablation study, we found that the proposed zero-shot task distribution and generative meta-learning is a key component for improving the model performance.

    		\subsection{Softmax and SVM} 
    		The proposed approach is generative, and so we can generate the samples of any class/distribution given the novel class attribute/description. Once we have the novel class synthetic data, we can train any traditional classifier for the unseen class data. Here we show the ZSL results of two standard classifiers, softmax, and linear-SVM, that are trained on the synthesized novel/unseen class samples. We find that both the classifiers have very competitive results and in some cases, linear-SVM shows a slightly better performance than softmax. We suggest that the reason behind this difference in performance is because linear-SVM learns the max-margin classifier while linear softmax classifier using neural ignores the max-margin. Refer to Table-[\ref{tab:zsl}] for the results of softmax and linear-SVM classifier.
    		\nocite{saligram2016learningJoint}

    \end{document}